\documentclass{article}

\usepackage[final]{corl_2019} 

\usepackage{graphics} 
\usepackage{graphicx}
\usepackage{epsfig} 
\usepackage{amsmath} 
\usepackage{amssymb}  
\usepackage{subfigure}
\usepackage{float}      
\usepackage{url}
\usepackage{multirow}
\usepackage{subfigure}
\usepackage{wrapfig}
\usepackage{tabularx}
\usepackage{tablefootnote}
\usepackage{hyperref}
\usepackage[usenames,dvipsnames]{xcolor}
\usepackage{bm}
\usepackage{array}
\usepackage{booktabs}
\usepackage{lipsum}
\usepackage{multicol}

\usepackage{accents} 
\usepackage[algoruled, resetcount, linesnumbered ]{algorithm2e}
\usepackage{algpseudocode}
\newlength{\mylen}
\let\oldnl\nl
\newcommand{\nonl}{\renewcommand{\nl}{\let\nl\oldnl}}
\SetKwProg{ParFor}{par for}{ :}{}
\SetKwProg{PyIf}{if}{ then:}{}
\SetKwProg{GetCorrelation}{Get Correlation}{ then:}{}

\graphicspath{{./figs/}}
\newcolumntype{C}[1]{>{\centering\let\newline\\\arraybackslash\hspace{0pt}}m{#1}}
\newcolumntype{Y}{>{\centering\arraybackslash}X}
\DeclareMathOperator*{\argmax}{arg\,max}
\DeclareMathOperator*{\argmin}{arg\,min}

\newcommand{\kk}{t}   
\newcommand{\pose}{\bm{p}}   

\newcommand{\power}{\bm{Z}}
\newcommand{\params}{\alpha}
\newcommand{\exponent}{\beta}
\newcommand{\likeparams}{\alpha}

\newcommand{\deltax}{\Delta x}
\newcommand{\deltay}{\Delta y}
\newcommand{\deltatheta}{\Delta \theta}
\newcommand{\transpose}{T}

\newcommand{\discross}{\star}

\newcommand{\cart}{\mathsf{C}}

\newcommand{\rotated}{\mathsf{R}}

\newcommand{\fft}{\mathsf{fft2d}}

\newcommand{\mask}{\bm{M}}
\newcommand{\rep}{\bm{S}}

\title{Masking by Moving: Learning Distraction-Free Radar Odometry from Pose Information}
\author{
  Dan Barnes, Rob Weston, Ingmar Posner\\
  Applied AI Lab,  University of Oxford \\
  \texttt{\{dbarnes, robw, ingmar\}@robots.ox.ac.uk} \\
}

\begin{document}
\maketitle


\begin{abstract}
This paper presents an end-to-end radar odometry system which delivers robust, real-time pose estimates based on a learned embedding space free of sensing artefacts and distractor objects. The system deploys a fully differentiable, correlation-based radar matching approach. This provides the same level of interpretability as established scan-matching methods and allows for a principled derivation of uncertainty estimates. The system is trained in a (self-)supervised way using only previously obtained pose information as a training signal. Using 280km of urban driving data, we demonstrate that our approach outperforms the previous state-of-the-art in radar odometry by reducing errors by up $68\%$ whilst running an order of magnitude faster. 
\end{abstract}


\section{Introduction}\label{sec:introduction}Robust ego-motion estimation and localisation are established cornerstones of autonomy. Emerging commercial needs as well as otherwise ambitious deployment scenarios require our robots to operate in ever more complex, unstructured environments and in conditions distinctly unfavourable for typical go-to sensors such as vision and lidar. Our robots now need to see further, through fog, rain and snow, despite lens flare or when directly facing the sun. Radar holds the promise of remedying many of these shortcomings. However, it is also a notoriously challenging sensing modality: radar applications are typically blighted by heterogeneous noise artefacts such as ghost objects, phase noise, speckle and saturation. In response, previous approaches to utilising radar for robot navigation have often tried to manually extract features from noise corrupted radar scans, commonly relying on simplifying assumptions on the distribution of power returns \citep{vivet2013localization}, manually designed heuristics \citep{cen2018precise}, or features designed for different modalities \citep{callmer2011radar, schuster2016landmark}. Nevertheless, the recent seminal work by Cen et al. \citep{cen2018precise} has firmly established radar as a feasible alternative to complement existing navigation approaches when it comes to ego-motion estimation. 

Beyond the basic methodology for pose estimation, the prevalence of vision- and lidar-based approaches in this space has given rise to a number of useful methods beyond those currently utilised for radar. State-of-the-art visual odometry, for example, leverages learnt feature representations \cite{wang2017deepvo} as well as attention masks filtering out potentially distracting objects \cite{barnes2018driven}. Lidar-based methods using correlative scan matching \cite{maddern2015leveraging} typically achieve highly accurate and intuitively interpretable results. 

\begin{figure}[]
  \centering
  \includegraphics[width=1.\columnwidth]{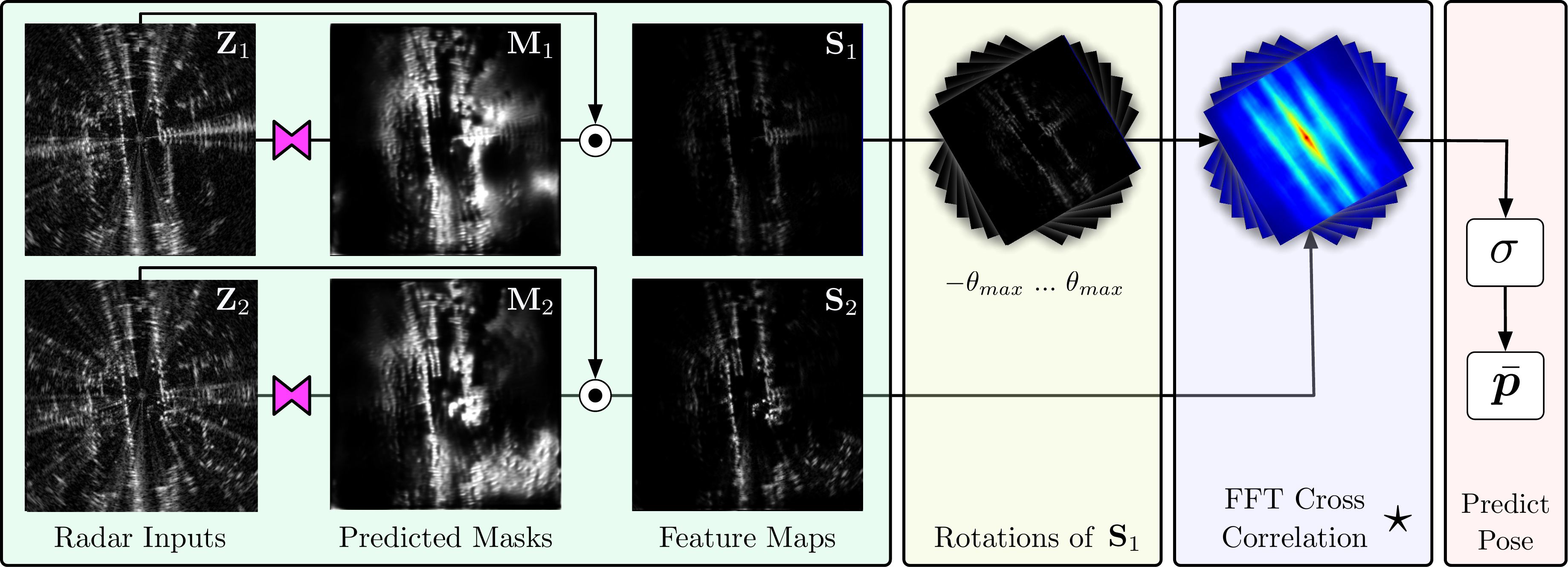}
\vspace{-4mm}
  \caption{
  Using masked correlative scan matching to find the optimum pose. Each radar scan is passed to a CNN (pink) in order to generate a mask which is subsequently applied to the input radar scan generating sensor artefact and distractor free representations $\rep_1$ and $\rep_{2}$. We then calculate the 2D correlation between $\rep_{2}$ and rotated copies of $\rep_1$ using the $\fft$ in order to generate the correlation volume $\bm{C}$. Finally we perform a softargmax operation in order to retrieve the predicted pose. Crucially this pipeline is fully  differentiable allowing us to learn the filter masks end to end. 
  A video summary of our approach can be found at: \url{https://youtu.be/eG4Q-j3_6dk}
  }
  \label{fig:hero-fig}
\end{figure}

Inspired by this prior art, the aim of our work is to provide a robust radar odometry system which is largely unencumbered by either the typical radar artefacts or by the presence of potentially distracting objects. Our system is explicitly designed to provide \emph{robust}, \emph{efficient} and \emph{interpretable} motion estimates. To achieve this we leverage a deep neural network to learn an essentially artefact and distraction free embedding space which is used to perform efficient correlative matching between consecutive radar scans. Our matching formulation is fully differentiable, allowing us to explicitly learn a representation suitable for accurate pose prediction. The correlative scan matching approach further allows our system to efficiently provide principled uncertainty estimates. 

Training our network on over 186,000 examples generated from 216km's of driving, we outperform the previous state of the art in challenging urban environments, reducing errors by over $68\%$ and running an order of magnitude faster. 
Furthermore, our pose ground truth is gathered in a self-supervised manner, automatically optimising odometry, loop closure, and location constraints, enabling us to adapt to new locations and sensor configurations with no manual labelling effort.

\section{Related Work}\label{sec:related-work}Compared to other sensing modalities such as vision or lidar, radar has received relatively little attention in the context of robot navigation. Prior art in this area largely deploys a more traditional processing pipeline consisting of separate feature extraction, data association and loss minimisation steps, for example using the Iterative Closest Point (ICP) algorithm \cite{besl1992method, ward2016vehicle}. For feature extraction some works deploy approaches developed in vision, such as SIFT and SURF \cite{callmer2011radar, schuster2016landmark}, others more bespoke methods such as CFAR filtering \cite{rohling2011ordered, vivet2013localization}, temporal-space continuity modelling \cite{jose2005augmented, jose2004relative}, and grid-map features such as Binary Annular Statistics Descriptor (BASD) \cite{rapp2016fscd}. Most recently the authors of \cite{cen2018precise} find point correspondences between point features extracted from raw scans using a shape similarity metric. The final pose is then found by minimising the mean squared error between point correspondences in close to real time.

By making use the Fourier transform correlation-based approaches are in contrast able to perform a dense search over possible point correspondences \cite{checchin2010radar} yielding intuitively interpretable results.
Similar approaches have also been applied successfully to lidar scan matching utilising efficient GPU implementations \cite{olson2009real} \cite{maddern2015leveraging}. In comparison to ICP, correlation-based methods have been shown to be significantly more robust to noise in pose initialisation \cite{olson2009real}. While robustness and interpretability are desirable, correlation-based methods operate on the assumption that the power returns from a particular location are stationary over time so that a correlation operation produces meaningful results. In reality, this is often not the case -- for example when dynamic objects are present in the scene. This problem is particularly pronounced in radar data due to the prevalence of noise artefacts.

Visual odometry systems, in contrast to radar-based ones, have a significant track record of successful application in robotics and beyond. While traditional processing pipelines similar to the one outlined for radar above have been widely deployed in this context (e.g. \cite{WinstonChurchill}) there has recently been significant interest in moving away from separate processing steps towards end-to-end approaches. Typically, a neural network is used to regress to a predicted pose directly from consecutive camera images, learning the relationship between features and point correspondences in an integrated manner (e.g. \cite{wang2017deepvo, detone2016deep}). In \cite{li2018undeepvo} the authors extend this approach by learning to predict the optimum pose from stereo images alone. As in many related fields, these end-to-end approaches demonstrate the potential for learning representations generally useful for odometry prediction. However, this comes at the expense of entangling feature representation and data association, which makes the resulting system significantly less interpretable. In contrast, the authors of \cite{barsan2018learning} propose to learn a feature embedding for localising online lidar sweeps into a previously known map, whilst maintaining the interpretability, of a conventional correlative scan matching approach.

Due to the ubiquitous nature of vision-based systems researchers have also addressed challenges beyond the basic pose estimation task such as suppressing noise sources inherent in individual scenes. For example, both \cite{mcmanus2013distraction} and \cite{barnes2018driven} try to mask areas of an image where non-stationary features might be found, which could corrupt the odometry estimate. Of particular relevance is \cite{barnes2018driven}, where a deep neural network is trained using data from other parts of the autonomous system in order to predict human interpretable ephemerality masks indicating the presence of distractor objects in a scene.

Given the large body of evidence that end-to-end approaches tend to outperform more traditional, hand-engineered processing pipelines it is tempting to conclude that our goal here is simply to deploy a deep network to radar odometry. And we do indeed leverage deep learning in our system. However, in doing so we are cognisant that we desire a system which ideally exploits the power of representation learning offered by end-to-end approaches while at the same time leveraging the efficiency, robustness and interpretability offered by correlation-based methods. Thus, inspired by \cite{barnes2018driven} and similar to \cite{barsan2018learning}, we deploy a correlation-based matching method as part of an end-to-end system which learns a radar embedding used to produce largely artefact and distraction-free representations optimised for pose prediction. Both the masks obtained as well as the cost-volumes considered remain as interpretable as more traditional approaches.

\section{Deep Correlative Scan Matching with Learnt Feature Embeddings}\label{sec:method}Given two consecutive radar observations $(\power_t, \power_{t-1})$ we wish to determine the relative pose $[\bm{R} | \bm{t}] \in \mathbb{SE}(2)$ giving the transformation between the two co-ordinate systems at each time step. In achieving this we aim to harness the efficiency, interpretability and robustness of correlative scan matching assuming that the power returned from each world location is independent of the co-ordinate system it was sensed in.
In reality the power returns generated from real world scenes are far from stationary, as dynamic objects move into and out of the field of view of the sensor and pertinent, random noise artefacts obscure the true power returns, limiting the performance of an out-of-the-box correlative scan matching system applied to radar data.

To address this, and inspired by the recent successes of learnt masking for pose prediction in vision \cite{barnes2018driven}, we instead perform correlative scan matching over a learnt feature embedding, utilising a deep, fully convolutional network to mask each radar scan as illustrated in Figure \ref{fig:hero-fig} (described in Section \ref{sec:correlative_scan_matching_with_learnt_features}). Through this approach we are able to harness the power of deep representation learning whilst ensuring the feature representation remains interpretable through the geometrical constraints imposed by the use of a correlative scan matching procedure. Crucially, we train our network by supervising pose prediction \emph{directly}. 
In doing so, our network naturally learns to attenuate distractor objects such as moving vehicles and sensor noise as they degrade pose estimation accuracy, whilst preserving features which are likely to be consistent between scans such as walls and buildings. This leads to a $68\%$ reduction in errors over the current state-of-the-art whilst, by making use of efficient correlation computations using the Fast Fourier Transform (FFT), running an order of magnitude faster.

Even in the limit of perfectly stationary power returns, uncertainty in our pose prediction still emanates from pathological solutions arising from the underlying scene topology. In Section \ref{sec:results_uncertainty_prediction} we show how we are additionally able to quantify the uncertainty in our pose prediction, further aiding the interpretability of our system.

\subsection{Correlative Scan Matching with Learnt Feature Embeddings}
\label{sec:correlative_scan_matching_with_learnt_features}
Let $(\power_{t-1}, \power_{t}) \in [0, 1]^{W\times H}$ denote consecutive observations made by single sweeps of the radar sensor, converted to Cartesian co-ordinates such that $\power^{u, v}_{t}$ gives the power return at Cartesian co-ordinate $(x, y)$ at time $t$. Let  $\pose = [\deltax, \deltay, \deltatheta]^\transpose$ denote the parameters of the relative pose $[\bm{R} | \bm{t}] \in \mathbb{SE}(2)$ between the co-ordinate frames at $t-1$ and $t$. We aim to predict the optimum pose from consecutive radar observations harnessing the efficiency, interpretability and robustness of correlative scan matching,
\begin{equation}
\label{eq:disc_cc_masked_opt}
    \bar{\pose} = \argmax_{\pose \in \mathbb{SE}(2)} \power_{\kk} \discross \power_{t -1}
\end{equation}
where $\power_{\kk} \discross \power_{\kk -1}$ is defined as the \emph{discrete cross correlation} between $\power_{\kk}$ (after being warped by the pose $\pose$) and $\power_{\kk -1}$.

In order to solve for the predicted pose $\bar{\pose}$ we consider a brute force approach: we discretise our search search space, calculating the cross correlation score for each pose on a regular grid of pose candidates before utilising a soft-argmax operation to solve for the optimum pose to sub-grid resolution accuracy. This is achieved efficiently using Algorithms \ref{alg:correlation_volume} and \ref{alg:soft_argmax}. By utilising bi-linear interpolation for all re-size and rotation operations, and computing the cross-correlation using the highly efficient 2D Fast Fourier Transform, we are able to search for the optimum pose over a large search area, efficiently solving (\ref{eq:disc_cc_masked_opt}) whilst still maintaining end-to-end differentiability.

Central to this approach is an assumption that the power returned from each world location is independent of the co-ordinate system it was sensed in. This assumption rarely holds in practice. Random noise artefacts, dynamic objects and changing scene occlusion cause fluctuations in the power field, degrading the accuracy of conventional correlation-based approaches applied to radar. To counter this, we propose to learn a feature representation $\rep$ specifically optimised for correlative scan matching by filtering each radar scan $\rep = \mask \odot \power$ with a mask $\mask = f_\alpha(\power)$ generated by a neural neural network $f_\params$ (where $\odot$ denotes Hadamard product). By limiting each element of the mask to $[0, 1]$ (using an element wise sigmoid), the network is able to learn to filter out distractor objects and noise in each sensor observation, before correlative scan matching is applied to find the optimum pose. By leveraging the differentiability of our approach for predicting $\bar{\pose}$, we are able to use Algorithm $\ref{alg:training}$, to learn a radar feature embedding specifically optimised for correlative scan matching by minimising the Mean Squared Error (MSE) over the training set, $\mathcal{D} = \{(\power_t, \power_{t-1}, \pose)^n\}_{n=1}^N$,
\begin{equation}
\label{eq:pose_supervision}
        \likeparams^* = \argmin_{\likeparams} \mathbb{E}_{\pose \sim \mathcal{D}}\big[|| \bar{\pose} -  \pose ||^2\big]
\end{equation}
to update our network parameters $\alpha$ using conventional stochastic gradient descent based optimisers.

\subsection{Pose Uncertainty Estimation}
\label{sec:pose_uncertainty}
Pathological solutions arising from the underlying scene topology increase the uncertainty in our pose prediction even in the case of perfectly stationary power returns. In the real world identifying such cases is important in order to ensure robust operation. To this end,
our approach also affords us a principled mechanism to estimate the uncertainty in each element of the predicted pose. 

In performing the soft-argmax operation, we first apply a temperature controlled softmax over the correlation scores for each candidate pose, to give weights $\bm{\omega} = \text{Softmax}(\exponent \bm{C})$, interpreted as the probability that each pose candidate is optimum. Assuming that our predicted pose is Gaussian distributed we can quantify the uncertainty in each pose prediction by using the weights $\bm{\omega}$ to predict both the mean pose $\bar{\pose}$ \emph{and} the predicted co-variance $\bar{\bm{\Sigma}}$,
\begin{gather}
\label{eq:gaussian_posterior}
    \bar{\pose} = \sum_s \omega_s \pose_s  \quad  \bar{\bm{\Sigma}} =  \sum_{s}  \omega_s \pose_s \pose_s^{\transpose} - \bar{\pose} \bar{\pose}^\transpose \quad  p(\pose | \rep_t) \approx \mathcal{N}(\pose | \bar{\pose}, \bar{\bm{\Sigma}})
\end{gather}
where we sum over all pose candidates. The softmax temperature parameter $\exponent$ plays an important role here: for high $\exponent$ our system is biased to the pose candidate with highest correlation and a low co-variance, whilst for low $\exponent$ to a weighted mean over a greater number of pose candidates and high co-variance.

\begin{minipage}{0.46\linewidth}
\centering
\begin{algorithm}[H]
\caption{Training} \label{alg:training}
\KwIn{}
\BlankLine
\nonl $\mathcal{D}$ \tcp{Dataset}
\nonl $\bm{r}$ \tcp{Search Region giving min and max range in $\deltax, \deltay, \deltatheta$}
\nonl $\bm{\delta}$ \tcp{Grid resolution in each dimension $\delta_x, \delta_y, \delta_{\theta}$}
\nonl $\exponent$ \tcp{Softmax Temperature Parameter}
\nonl $\epsilon$ \tcp{Learning Rate}
\nonl $\alpha$ \tcp{Initial Network Parameters}

\BlankLine
$\bm{G}_{x y\theta} = MeshGrid(\bm{r}, \bm{\delta})$ \\
\BlankLine
\SetAlgoLined
\While{not converged}{
    $\bm{Z}_1, \bm{Z}_2, \pose \leftarrow Sample(\mathcal{D}$) \\
    \BlankLine
    $\mask_1, \mask_2 \leftarrow f_\alpha(\bm{Z}_1), f_\alpha(\bm{Z}_2$) \\
    $\rep_1, \rep_2 \leftarrow \mask_1 \odot \bm{Z}_1, \mask_2 \odot \bm{Z}_2$ \\
    \BlankLine
    $\bm{C} \leftarrow GetCorrelation (\bm{G}_{xy\theta}$, $\rep_1$, $\rep_2)$ \\
    $\bar{\pose} \leftarrow SoftArgMax(\bm{G}_{xy\theta}, \bm{C}, \beta)$ \\
    \BlankLine
    $\likeparams \leftarrow \likeparams - \epsilon \nabla_{\likeparams} \mathcal{L}(\bar{\pose}; \pose)$ \\
}
\end{algorithm}
\end{minipage} \hspace{5mm}%
\begin{minipage}{0.46\linewidth}
\centering
\begin{algorithm}[H]
\caption{Correlation} \label{alg:correlation_volume}
  \SetKwProg{Fn}{function}{ :}{}
  \Fn{GetCorrelation($\bm{G}_{xy\theta}$, $\bm{X}_1$, $\bm{X}_2)$}{
  $n_x, n_y, n_\theta \leftarrow Shape(\bm{G}_{xy\theta})$ \\
  $\bm{C} = Zeros([n_x, n_y, n_\theta])$ \\
  $\bm{G}_{xy}, \bm{G}_\theta \leftarrow \bm{G}_{xy\theta}$ \\
  $\bm{X}_1, \bm{X}_2 \leftarrow Resize(\bm{X}_1, \bm{X}_2, \bm{G}_{xy})$
  \\
  \ParFor{$i \leftarrow 1$ \KwTo $n_\theta$}{
  $\bm{X}_1^{\rotated} \leftarrow Rotate(\bm{X}_1, \bm{G}_\theta[i])$ \\
  $\bm{C}[:, :, i] \leftarrow \fft^{-1} \big( \fft(\bm{X}_1^{\rotated}) \odot \fft(\bm{X}_2^{\cart})\big)$ \\
  }
  \Return{$\bm{C}$}}
  \end{algorithm}
  \begin{algorithm}[H]
  \caption{Soft Arg Max} \label{alg:soft_argmax}
  \SetKwProg{Fn}{function}{ :}{}
  \Fn{SoftArgMax($\bm{G}_{xy\theta}$, $\bm{C}$, $\beta$)}{
    $\bm{\omega} \leftarrow Softmax(\exponent \bm{C})$ \\
    $\bm{G}_x, \bm{G}_y, \bm{G}_\theta \leftarrow \bm{G}_{xy\theta}$ \\
    $\deltax \leftarrow \sum_{i, j, k} \big( \bm{\omega} \odot \bm{G}_x \big)[i, j, k]$ \\
    $\deltay \leftarrow \sum_{i, j, k} \big( \bm{\omega} \odot \bm{G}_y \big)[i, j, k]$ \\
    $\deltatheta \rightarrow \sum_{i, j, k} \big( \bm{\omega} \odot \bm{G}_\theta \big)[i, j, k]$ \\
    $\Return [\deltax, \deltay, \deltatheta]$ \\
  }
\end{algorithm}
\end{minipage}

\section{Experimental Setup}\label{sec:experimental-setup}\subsection{Dataset}\label{sec:experimental-dataset}
To evaluate our approach we use the recently released Oxford Radar RobotCar Dataset \cite{RadarRobotCarDatasetArXiv}, a radar extension to the Oxford RobotCar Datsset \cite{RobotCarDatasetIJRR}, which provides Navtech CTS350-X radar data as well as ground truth poses. The Navtech CTS350-X is a Frequency Modulated Continuous Wave (FMCW) scanning radar without doppler information, configured to return 3768 power readings at a resolution of $4.32$cm across 400 azimuths at a frequency of $4$Hz (corresponding to a maximum range of 163m). The beam spread is 2 degrees in azimuth and 25 degrees in elevation with a cosec squared beam pattern. We randomly split the traversals into training ($80 \%$) and evaluation ($20\%$) partitions. We additionally run spatial cross validation experiments, where each split occupies a different real world region of the dataset. Further information on these results and the dataset can be found in the appendix \ref{sec:dataset_splits}, \ref{sec:cross-validation-results}.

To validate the advantages of learning masks directly from pose supervision we compare against supervising the learnt masks directly on the proxy task of predicting temporally static occupied cells. Training data for this is generated using a similar approach to \cite{barnes2018driven}. For each radar scan we warp the nearest radar sensor observation from each training traversal into the current pose before applying a static power threshold. We then form a 2D histogram counting the number of thresholded power returns that fall in each Cartesian grid cell. Any grid cell with more than 9 consistent observations is assumed to be temporally stable and is labelled with a 1, whilst every other cell is set to 0. This is repeated for every pose in every dataset. Examples of the masks generated by this approach can be found in the appendix \ref{sec:baseline-masks}.

\subsection{Network Architecture and Training}\label{sec:experimental-network}
In all experiments we use a U-Net style architecture \cite{ronneberger2015u} in which we encode the input tensor through the repeated application of two convolutional layers (filter size 3x3) with ReLU activations before a max pooling operation. After each max pool the width and height of the tensor are reduced by a factor of 2 whilst the number of features is doubled, starting from 8 at the input to 256 at the bottleneck of the network (corresponding to 5 max pools). The feature tensor is then converted back to the original shape by the decoder through the application of bilinear upsampling followed by two convolutional layers increasing the width and height and decreasing the feature channels by a factor of 2. Skip connections at each level are implemented allowing information to flow from encoder to decoder by stacking each representation with the output from the bilinear upsampling layer in each case. The final convolutional layer has a single output channel with a sigmoid activation to limit the range to $[0,1]$. We experiment with learning to mask both Cartesian and Polar radar representations, as well as both \emph{single} and \emph{dual} configurations. In the dual case radar observations are concatenated and passed as a single input producing two masks (instead of one) at the output. An architecture diagram can be found in the appendix \ref{sec:network-architecture}. In all cases we consider a search region of $[-50\text{m}, 50\text{m}]$ in $\deltax$ and $\deltay$ and $[-\pi / 12, \pi/ 12]$ in $\deltatheta$. We experiment with the three grid resolutions $[0.2\text{m}, 0.4\text{m}, 0.8\text{m}]$ for $\delta_x$ and $\delta_y$ whilst fixing $\delta_{\theta}$ to $\pi / 360$.

Our network is implemented in Tensorflow \cite{abadi2016tensorflow} and trained using the Adam Optimiser \cite{kingma2014adam} \mbox{(learning} rate $1\text{e}-5$ and batch size 5) until the loss on a small validation set is a minimum. When training our network with pose supervision we minimise the loss proposed in (\ref{eq:pose_supervision}). We performed a grid search over the optimum value of $\beta$ and found setting it to 1 gave good performance. 

\subsection{Evaluation Metrics and Baselines}\label{sec:experimental-metrics}
\label{sed:uncertainty_evaluation}
Our primary baseline is the current state of art for radar odometry \cite{cen2018precise} (implemented in C++) in which the authors extract point features from consecutive radar scans before scan matching using a global shape similarity score and refining by minimising mean squared error. 
Our radar was set to a range resolution of $4.32$cm, whilst the original algorithm was developed for a $17.28$cm resolution. As such we compare against \cite{cen2018precise} with full resolution radar scans and downsampled (with max pooling) to $17.28$cm. For context we also provide visual odometry estimates (as in \cite{cen2018precise}). To assess the benefits of learning feature masks specifically optimised for pose prediction, we benchmark against scan matching on the raw radar scans without masking, as well as using the method proposed in \cite{barnes2018driven} with mask labels generated as described in Section \ref{sec:experimental-dataset}. In this setup, we supervise (using a binary cross entropy loss) the learnt masks directly (instead of supervising pose prediction). We also benchmark against taking an off the shelf deep odometry model and training this for the task of radar pose prediction. Specifically we use the UnDeepVO model proposed in \cite{li2018undeepvo}.

For all evaluations we follow the KITTI odometry benchmark \cite{geiger2012we}. For each 100m offset up to 800m, we calculate the average residual translational and angular error for every example in the datastet normalising by the distance travelled. Finally, we average these values. Due to highly skewed error distributions we report Inter Quartile Range (IQR) for each method instead of the standard deviation. All timing statistics are calculated using a 2.7 GHz 12-Core Intel Xeon E5 CPU and Nvidia Titan Xp GPU by averaging across 1000 predictions.

\subsection{Uncertainty Evaluation}
\label{sec:uncertainty_evaluation}
To assess the quality of the uncertainty predicted by our approach we observe that if our pose distribution is Gaussian than the Mahalanobis error
\begin{equation}
    d^2 = (\pose - \bar{\pose})^\transpose \bar{\bm{\Sigma}}^{-1} (\pose - \bar{\pose}),  \quad d^2 \sim \chi^2(3)
\end{equation}{}
\noindent should be chi-squared distributed with degrees of freedom equal to the state dimensionality of $\pose$ (in this case three). As the mean of a chi-squared distribution is equal to the distributions degrees of freedom, by averaging the mean Mahalanobis distance over the test dataset $\bar{d}^2 = \frac{1}{N} \sum_{n} d^2_n$ we can assess to what degree the uncertainties predicted by our approach are calibrated to the test errors \cite{maddern2015leveraging}. Specifically, if $\bar{d}^2 \ll 3$ then our model is overly conservative in its predictions whilst if $\bar{d}^2 \gg 3$ it is overly confident. In Section \ref{sec:results_uncertainty_prediction} we use this result to tune the temperature parameter $\exponent$ to provide us with realistic uncertainties, that are calibrated to the true errors in our system.

\section{Results}\label{sec:experiments}
In this section we evaluate the performance of our approach. We find by utilising correlative scan matching in combination with a learnt radar feature embedding we are able to significantly outperform the previous state of art in both prediction performance and speed. Additionally, we show how, by tuning the temperature parameter of the softargmax, we are able to predict realistic and calibrated uncertainties, further increasing the interpretability of our system and allowing us to identify pathological cases, crucial for robust operation in the real world.

\begin{table}[t!]
\centering
\label{table:results}
\small\addtolength{\tabcolsep}{-3.5pt}

\begin{tabularx}{\textwidth}{ lccccccc }
\toprule
                 & Resolution                                  & \multicolumn{2}{c}{Translational error ($\%$)} & \multicolumn{2}{c}{Rotational error (deg/m)} & \multicolumn{2}{c}{Runtime (s)}  \\ 
 \textbf{Benchmarks}   & (m/pixel) & Mean & IQR &  Mean & IQR &  Mean & Std.  \\ \cmidrule(lr){1-1} \cmidrule(lr){2-2} \cmidrule(lr){3-4} \cmidrule(lr){5-6} \cmidrule(lr){7-8}
RO Cen Full Resolution \cite{cen2018precise}      & 0.0432    & 8.4730            & 5.7873    & 0.0236            & 0.0181 & \textit{0.3059} & 0.0218 \\ \vspace{1mm}
RO Cen Equiv. Resolution \cite{cen2018precise}        & 0.1752    & \textit{3.7168}   & 3.4190    & \textit{0.0095}   & 0.0095 & 2.9036 & 0.5263 \\ \vspace{1mm}
Raw Scan                                    & 0.2       & 8.3778            & 7.9921    & 0.0271            & 0.0274 & 0.0886 & 0.0006 \\ \vspace{1mm}
Supervised Masks Polar                      & 0.2       & 5.9285            & 5.6822    & 0.0194           & 0.0197 & 0.0593 & 0.0014 \\ \vspace{1mm}
Supervised Masks Cart                       & 0.2       & 5.4827            & 5.2725    & 0.0180           & 0.0186 & 0.0485 & 0.0013 \\  \vspace{1mm}
Adapted Deep VO Cart \cite{li2018undeepvo}  & 0.2       & 4.7683            & 3.9256    & 0.0141           & 0.0128 & 0.0060 & 0.0003 \\ \vspace{1mm}
Adapted Deep VO Polar \cite{li2018undeepvo} &  -        & 9.3228            & 8.3112    & 0.0293            & 0.0277 & 0.0093 & 0.0002 \\ \vspace{2mm} 
Visual Odometry \cite{WinstonChurchill}     & -         & 3.9802            & 2.2324    & 0.0102            & 0.0065 & 0.0062 & 0.0003 \\

\textbf{Ours}                               &           &                   &           &                   &        &        &        \\  \cmidrule(lr){1-1}
Polar                                       & 0.8       & 2.4960   & 2.1108    & 0.0068   & 0.0052 & 0.0222 & 0.0013 \\
                                            & 0.4       & 1.6486   & 1.3546    & 0.0044   & 0.0033 & 0.0294 & 0.0012 \\  \vspace{1mm}
                                            & 0.2       & 1.3634   & 1.1434    & 0.0036   & 0.0027 & 0.0593 & 0.0014 \\ 
Cartesian                                   & 0.8       & 2.4044   & 2.0872    & 0.0065   & 0.0047 & 0.0113 & 0.0012 \\
                                            & 0.4       & 1.5893   & 1.3059    & 0.0044   & 0.0035 & 0.0169 & 0.0012 \\  \vspace{1mm}
                                            & 0.2       & 1.1721   & 0.9420    & 0.0031   & 0.0022 & 0.0485 & 0.0013 \\ 
Dual Polar                                  & 0.8       & 2.5762   & 2.0686    & 0.0072   & 0.0055 & 0.0121 & 0.0003 \\
                                            & 0.4       & 2.1604   & 1.9600    & 0.0067   & 0.0053 & 0.0253 & 0.0006 \\  \vspace{1mm}
                                            & 0.2       & 1.2621   & 1.1075    & 0.0036   & 0.0029 & 0.0785 & 0.0007 \\
Dual Cart                                   & 0.8       & 2.7008   & 2.2430    & 0.0076   & 0.0054 & \textbf{0.0088} & 0.0007 \\
                                            & 0.4       & 1.7979   & 1.4921    & 0.0047   & 0.0036 & 0.0194 & 0.0010 \\  \vspace{1mm}
                                            & 0.2       & \textbf{1.1627} & 0.9693    & \textbf{0.0030}   & 0.0030 & 0.0747 & 0.0005 \\
\bottomrule \\
\end{tabularx}

\caption{Odometry estimation and timing results. Here ``RO Cen'' \cite{cen2018precise} is our primary benchmark reported for $0.04\text{m}$ (full resolution) and, by downsampling, $0.17 \text{m}$ (equivalent resolution for which the approach was originally developed). For comparison we also provide performance results for correlative scan matching on the \emph{raw} power returns, for mask supervision (instead of supervising the predicted pose directly), and adapting the deep VO network proposed in \cite{li2018undeepvo}, alongside visual odometry \cite{WinstonChurchill} for context. All baselines performed best at 0.2 m/pixel resolution where applicable and the rest are omitted for clarity. We experiment with both polar and Cartesian network inputs at multiple resolutions. 
Our approach outperforms the current state of the art, ``RO Cen'' (italics), for all configurations of Cartesian / polar inputs and independent / dual masking at all resolutions.
Our best performing models in terms of speed and odometry performance are marked in bold.
}

\label{table:odometry-estimation}

\end{table}

\begin{figure}[h]
  \includegraphics[width=1.0\textwidth]{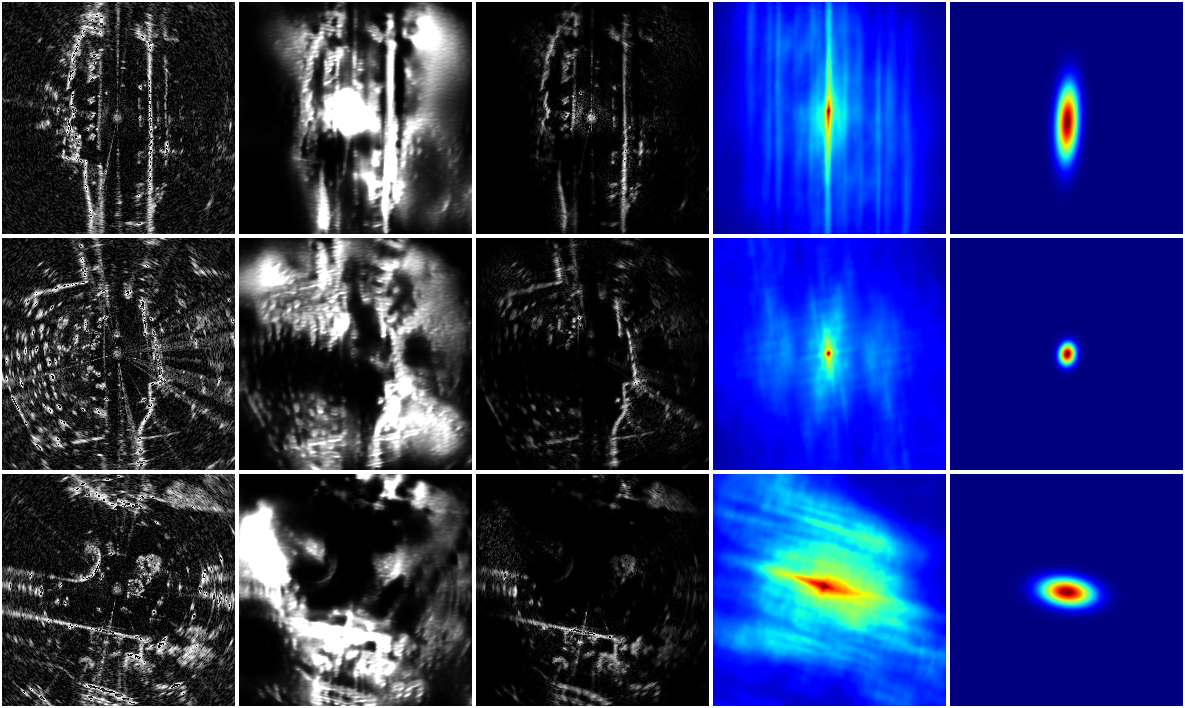}
  \begin{small}
  \begin{tabularx}{\textwidth}{YYYYY}
    Radar Input & Predicted Mask & Masked Features & Correlation & Covariance
  \end{tabularx}
  \end{small}
  \vspace{-3mm}
  \caption{
Qualitative examples generated from our best performing model. Our network learns to mask out noise and distractor objects whilst preserving temporally consistent features such as walls, well suited for pose prediction. Predicted co-variance is high for pathological solutions arising through a lack of constraints in the x-direction (top), whilst stationary well-constrained scenes result in low co-variance (middle). Motion blur increases the uncertainty due to ambiguous point correspondence (bottom). Further examples can be found in Figure \ref{fig:further_examples} in the appendix.}
  \label{fig:qualitative-results-allcart}
\end{figure}

\subsection{Odometry Performance}\label{sec:method-baseline}
Table \ref{table:odometry-estimation} gives our prediction and timing results. We experiment with both Cartesian and Polar inputs to the masking network (converting the latter to Cartesian co-ordinates before correlative scan matching), as well as experimenting with single and dual configurations as detailed in Section \ref{sec:experimental-network}.

At all resolutions and configurations we beat the current state of the art with our best model reducing errors by $68 \%$ in both translation and rotation, whilst running over 4 times faster. Our fastest performing model runs at over $100 \text{Hz}$ whilst still reducing errors on the state of the art by $28 \%$ in translational and $20 \%$ in rotational error (further results exploring the accuracy-speed trade off can be found in \ref{sec:speed-vs-accuracy}). 
We find that Cartesian network inputs typically outperform Polar (presumably because correlative scan matching is performed in Cartesian space). Dual input configurations also typically outperform passing single sensor observations to the masking network.

Key to our approach is learning a radar feature embedding that is optimised for pose prediction: compared to correlative scan matching on the raw radar power returns this allows us to reduce errors by over $85 \%$.  As predicted, optimising masks directly for pose prediction results in a higher prediction accuracy than mask supervision labelling the temporally stationary scene directly. We also find that simply adapting a deep odometry approach to radar results in significantly worse performance. Our approach in contrast makes use of the inherent top down representation of a radar observation which lends itself to a correlative scan matching procedure, whilst learning to mask out noise artefacts which make pose prediction in radar uniquely challenging. In addition, by adopting a correlative scan matching approach, our results remain interpretable: Figure \ref{fig:qualitative-results-allcart} shows several qualitative examples in which the network learns to mask noise artefacts and dynamic objects in the scene whilst preserving features which are likely to be temporally stationary such as walls.

\subsection{Uncertainty Prediction}
\label{sec:results_uncertainty_prediction}
In addition to the boosts in performance and speed afforded by our approach, we are also able to estimate the uncertainty in each pose prediction: by interpreting the weights generated through the temperature controlled softargmax operation as the probability that each pose candidate is optimum we predict the co-variance $\bar{\bm{\Sigma}}$ in our prediction as detailed in Section \ref{sec:pose_uncertainty}.

\begin{figure}[h]
    \label{fig:uncertainty-prediction}
    \centering
    \includegraphics[width=\linewidth]{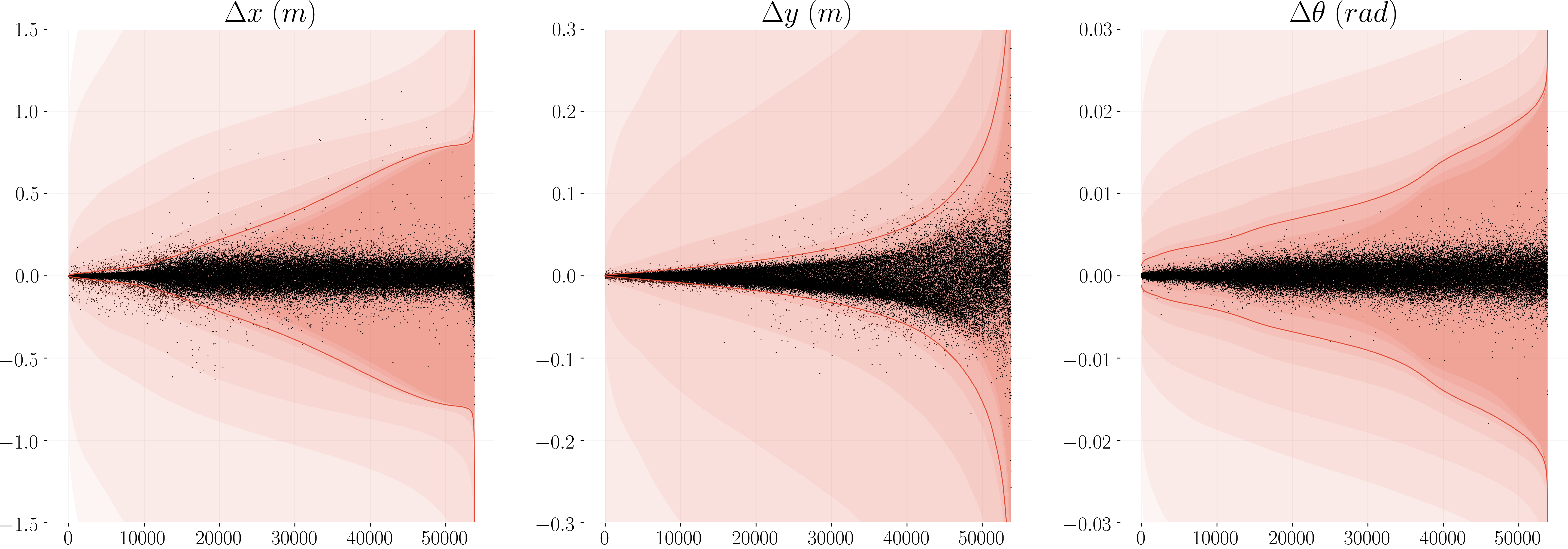}
    \vspace{-5mm}
    \begin{table}[H]
        \label{tab:mal_distance}
        \centering
        \begin{tabular}{@{}lccccccccc@{}}
            \toprule
            Temperature Parameter $\beta$           & 4.467    & 3.548  & \textbf{2.789} & 2.512 & 1.995 & 1.413 & 1.122 \\
            Mean Mahalanobis Distance $\bar{d}^2$ & 4268.471 & 57.099 & \textbf{2.992} & 1.244 & 0.283 & 0.065 & 0.030 \\ \bottomrule
        \end{tabular}
    \end{table}
    \vspace{-5mm}
    \caption{The marginal distributions and errors (black) in each pose component for each example in our test set ordered by predicted uncertainty. The colours correspond to 1.98 standard deviation bounds plotted for each of the temperature parameters given in the table with dark to light moving through the table left to right. The red line corresponds to the standard deviation bound plotted for $\beta = 2.789$ corresponding to a mean Mahalanobis distance of $\bar{d}^2 = 2.99$. For this temperature setting the majority of the errors fall within the $1.98$ standard deviation bound. Note the $y$ axis in each case has a different scale.}
\end{figure}

We now use the methodology proposed in Section \ref{sec:uncertainty_evaluation} to tune the temperature parameter $\exponent$ such that the mean Mahalanobis distance $\bar{d}^2 \approx 3$ producing uncertainties $\bar{\bm{\Sigma}}$ that are calibrated to the errors in our system. Naively perturbing the temperature parameter away from its original value $\beta_0$ degrades pose prediction performance as the feature mask no longer corresponds to the $\beta$ it was optimised for.
Instead, we calculate the predicted pose using $\exponent_0$, whilst varying $\exponent$ to tune the co-variance matrix. 
The results of this process (for the $0.8$m resolution single mode Cartesian model from Table \ref{table:odometry-estimation}) are shown in Figure \ref{fig:uncertainty-prediction} alongside the marginal distributions for the uncertainty in each pose component plotted with the true errors in our system ordered by predicted uncertainty. For a temperature parameter $\beta = 2.789$ the mean Mahalanobis distance $\bar{d}^2$ is equal to $2.99$ giving us well calibrated uncertainty predictions, whilst temperature parameters above and below this value are overly certain and conservative respectively. There is a clear correlation between error and uncertainty with most errors falling within the predicted uncertainty bounds.

 Figure \ref{fig:qualitative-results-allcart} shows Gaussian heat maps generated through our approach; the results are highly intuitive with feature embeddings well constrained in each dimension having smaller and symmetric co-variance, whilst pathological solutions arising from a lack of scene constraints increase the uncertainty in $\deltax$.
\vspace{-3mm}

\section{Conclusions}\label{sec:conclusions}By using a learnt radar feature embedding in combination with a correlative scan matching approach we are able to improve over the previous state of the art, reducing errors in odometry prediction by over $68\%$ and running an order of magnitude faster, whilst remaining as interpretable as a conventional scan matching approach. Additionally, our method affords us a principled mechanism by which to estimate the uncertainty in the pose prediction, crucial for robust real world operation.

Our approach for attaining calibrated uncertainties currently relies on tuning a pre-trained model. An interesting direction for future work would be to incorporate this tuning process into the training pipeline, learning not only a radar feature embedding optimised for pose prediction but also for uncertainty estimation. We leave this for future work.





\clearpage
\acknowledgments{This work was supported by the UK EPSRC Doctoral Training Partnership and EPSRC Programme Grant (EP/M019918/1). The authors also would like to acknowledge the use of Hartree Centre resources and the University of Oxford Advanced Research Computing (ARC) facility in carrying out this work (http://dx.doi.org/10.5281/zenodo.22558).}

\bibliography{rss2019}  

\newpage
\appendix 
\section{Implementation}

\subsection{Masking Network Architecture}\label{sec:network-architecture}
\begin{figure}[H]
  \centering
  \includegraphics[width=0.92\linewidth]{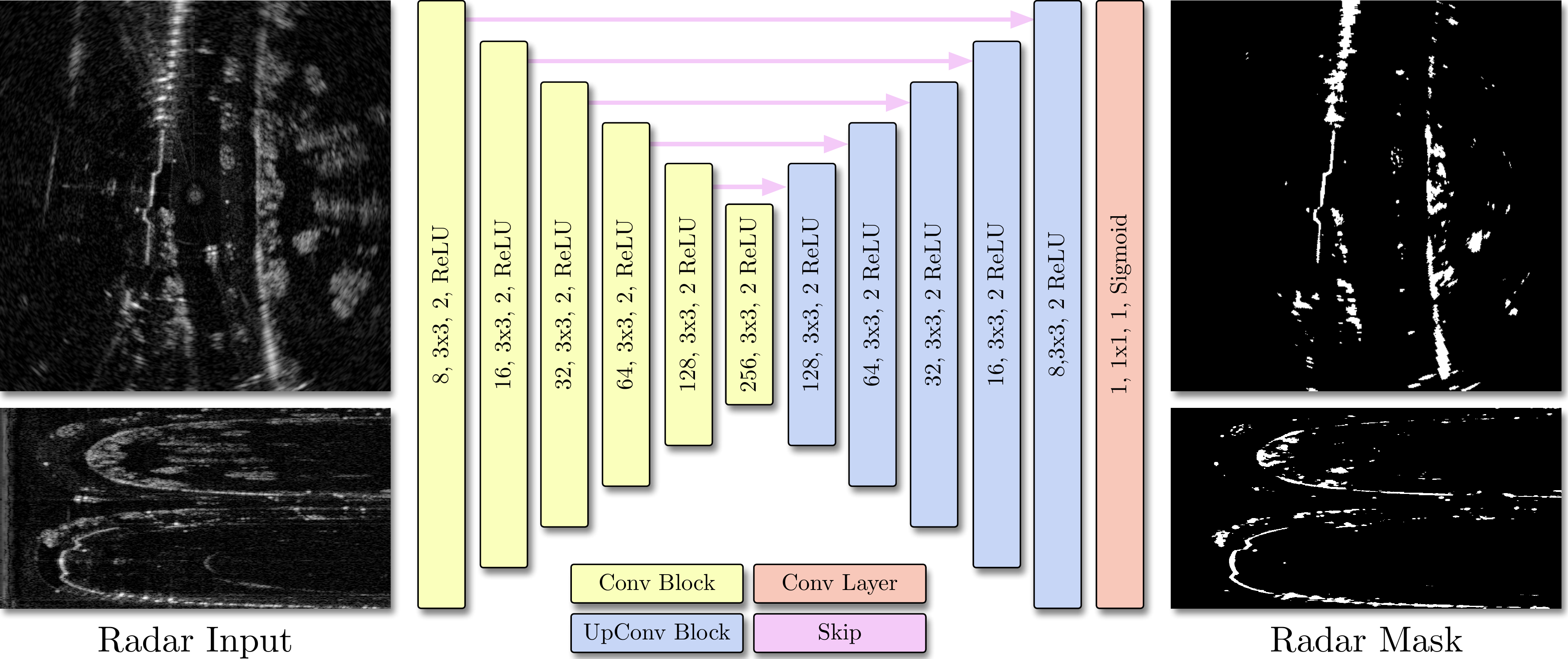}
  \caption{Architecture diagram of the radar masking network. Layers are detailed by output channels, kernel sizes, repetitions and activations respectively. The final network layer has a single output channel with a sigmoid activation to limit the masking range to $[0,1]$. We experiment using the masking network in both Cartesian and Polar radar representations. Additionally we investigate the impact of modifying the \emph{single} configuration shown to \emph{dual} configuration, in which sequential radar observations used for odometry prediction are concatenated and passed as a single input producing two masks (instead of one) at the output. For more details please refer to the text in Section \ref{sec:experimental-network}. The predictions shown are from a network directly supervised with baseline masks detailed in Section \ref{sec:baseline-masks}.}
  \label{fig:network-architecture}
\end{figure}


\subsection{Speed vs Accuracy Trade Off}\label{sec:speed-vs-accuracy}
\begin{figure}[H]
  \centering
  \includegraphics[width=0.475\linewidth]{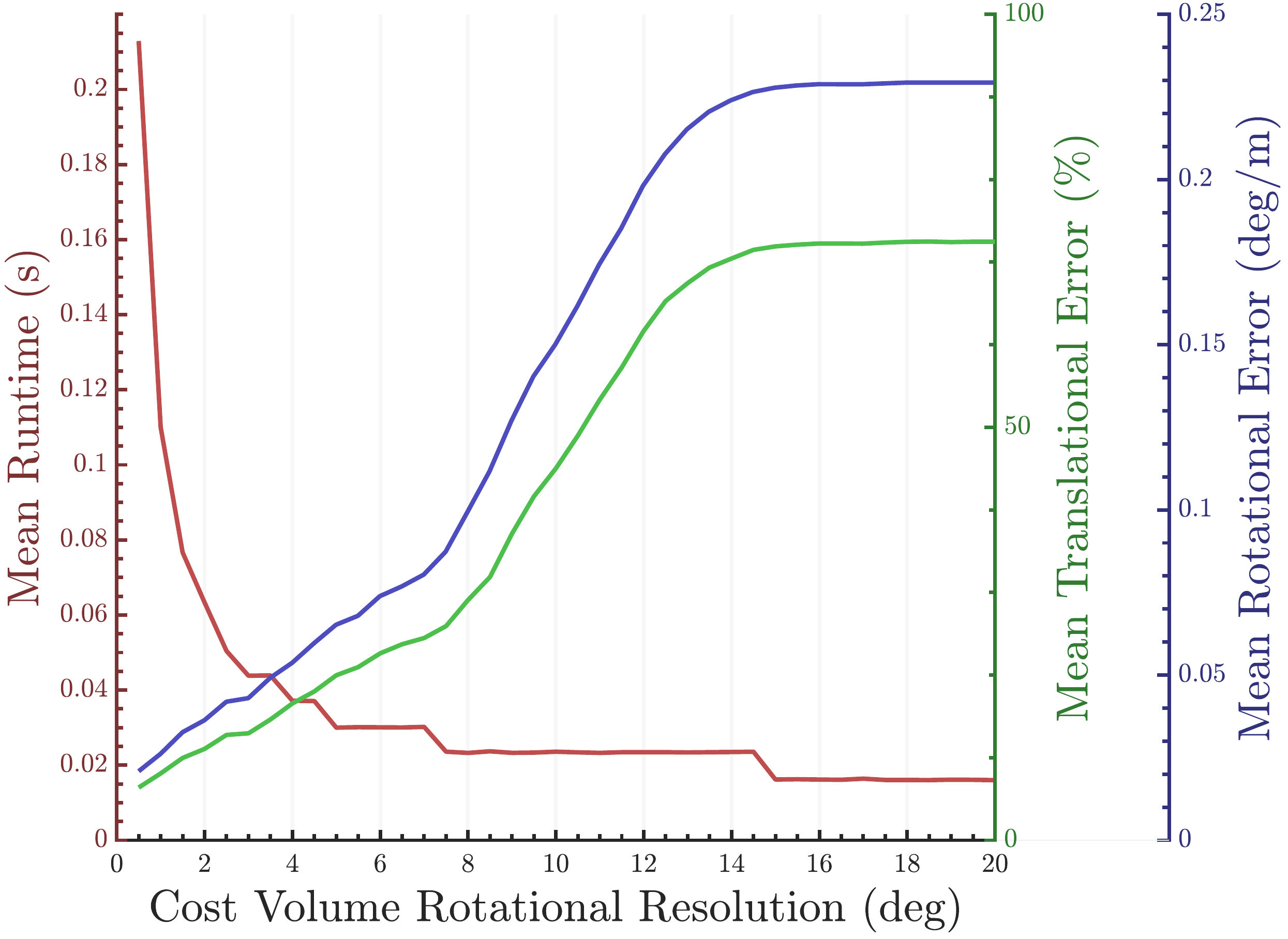}
  \hfill
  \includegraphics[width=0.475\linewidth]{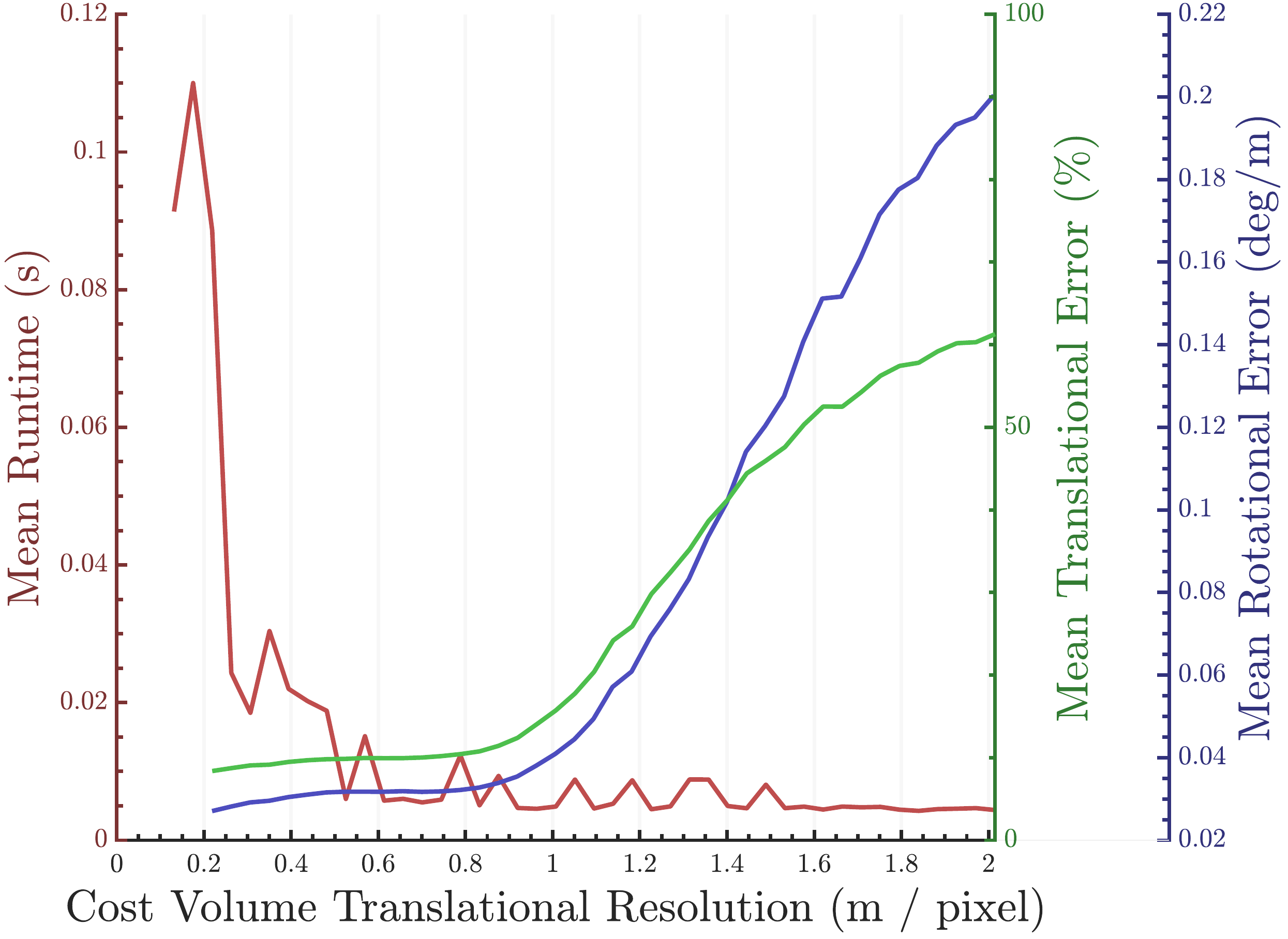}
  \caption{Translational error (green), angular error (blue) and run time (red) as a function of Cost volume resolution in degrees (left) and metres per pixel (right). In the case of limited computational resources or required pose estimate accuracy it is possible to flexibly trade off performance and computational speed.}
  \label{fig:results-xcorr-baseline-sweep}
\end{figure}

By reducing our Cartesian grid resolution before calculating the correlation volume, for the same grid coverage we are able to predict the optimum pose in a shorter amount of time to the detriment of pose prediction accuracy. Estimating this trade off for our trained models is challenging and requires many training runs. Instead we investigate the speed-accuracy trade off by performing correlative scan matching on the raw power returns at a variety of grid resolutions according to Algorithms \ref{alg:correlation_volume} and \ref{alg:soft_argmax}. The results for this process are displayed in Figure \ref{fig:results-xcorr-baseline-sweep} which we use to choose the grid resolutions for the main results presented in Table \ref{table:results}.

\section{Data}
\subsection{Baseline Masks}\label{sec:baseline-masks}
\sloppy
To validate the advantages of learning masks directly from pose supervision we compare against supervising the learnt masks directly on the proxy task of predicting temporally static occupied cells. To generate static mask labels we use a similar approach to \cite{barnes2018driven} as detailed in Section \ref{sec:experimental-dataset}, whereby nearby radar scans from different traversals are warped into the current sensor frame to assess temporal stability. Even with a large corpus of accurately labelled masks identifying static structure suitable for estimating odometry, we observe increased performance by training directly on the task of pose estimation.

\begin{figure}[H]
    \includegraphics[width=0.245\linewidth]{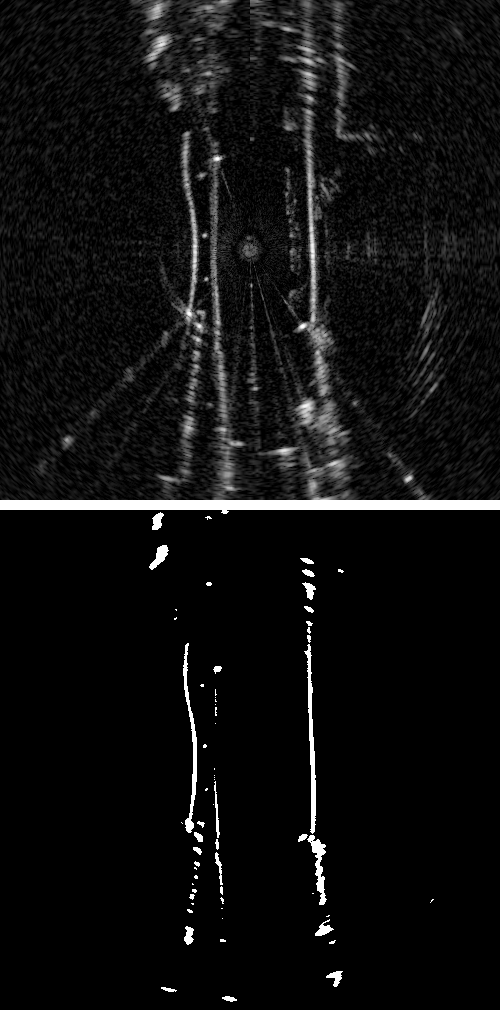}
    \hfill
    \includegraphics[width=0.245\linewidth]{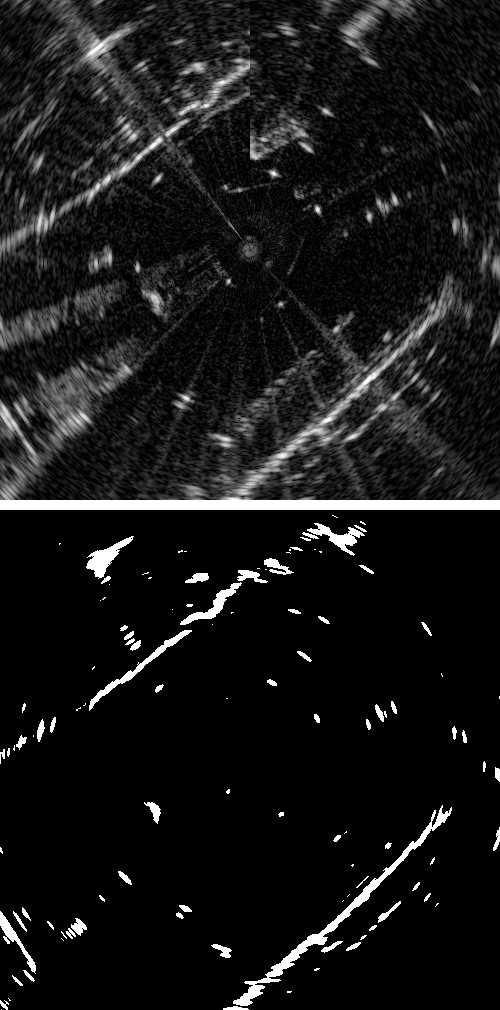}
    \hfill
    \includegraphics[width=0.245\linewidth]{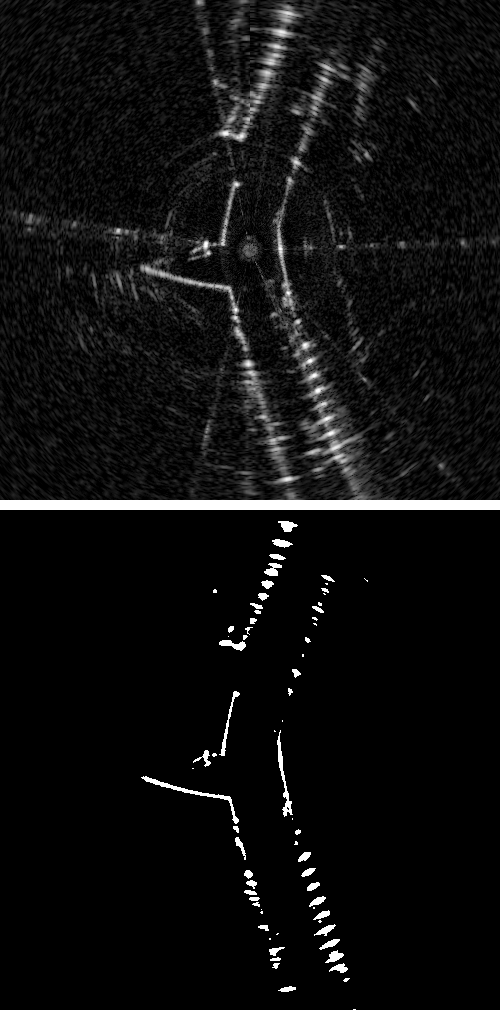}
    \hfill
    \includegraphics[width=0.245\linewidth]{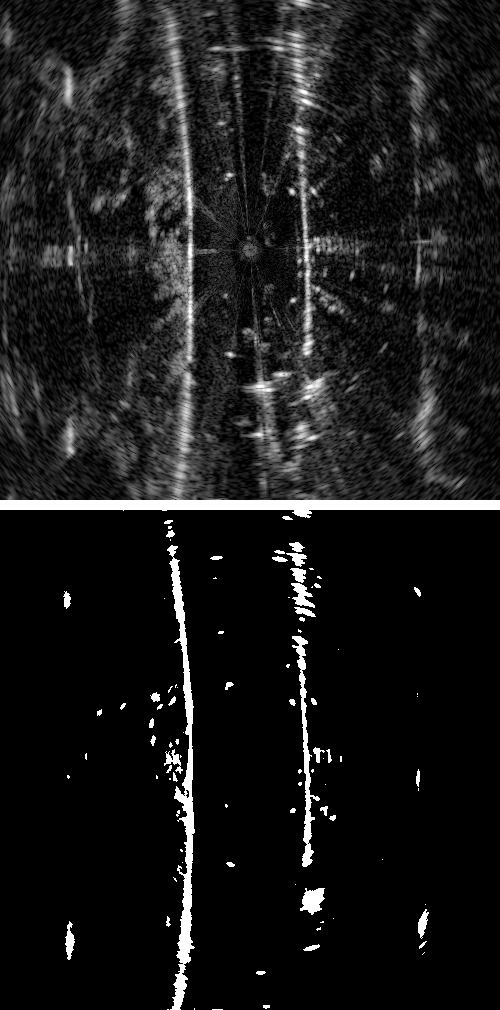}
  \caption{Example generated baseline masks used to supervise the radar masking network directly. For a given raw radar scan at time $t$ (top) we can automatically generate high quality baseline masks identifying structure useful for pose estimation (bottom).}
  \label{fig:baseline-masks}
\end{figure}


\subsection{Dataset Splits}\label{sec:dataset_splits}
\begin{figure}[H]
  \centering
  \includegraphics[width=0.95\linewidth]{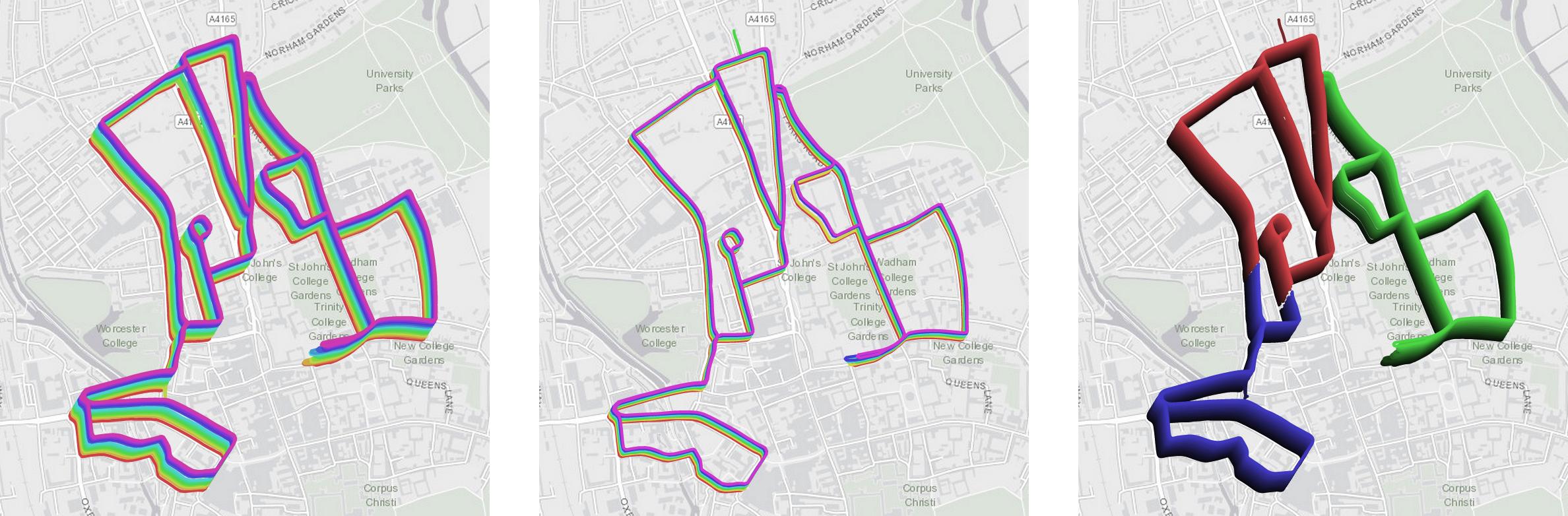}
  \begin{small}
  \begin{tabularx}{\textwidth}{YYY}
    Training Traversals & Testing Traversals & Spatial Cross Validation
  \end{tabularx}
  \end{small}
  \caption{Trajectories of the ground truth optimised pose chains used for the 25 training (left) and 7 evaluation (middle) traversals from the Oxford Radar RobotCar Dataset \cite{RadarRobotCarDatasetArXiv}  covering a wide variety of traffic and other challenging conditions in complex urban environments. In addition to splitting the dataset temporally we provide spatial cross validation results (right), detailed in Section \ref{sec:cross-validation-results}. Each traversal is incrementally offset with a unique colour for visualisation.}
  \label{fig:dataset_splits}
\end{figure}

\newpage

\section{Results}

\subsection{Spatial Cross Validation}\label{sec:cross-validation-results}
In Section \ref{sec:method-baseline} we achieve radar odometry performance far exceeding the state of the art.
However we train and evaluate on scenes from the same spatial locations. To assess how well our models generalise to previously un-seen scenes, in this section we train and evaluate our models using spatial cross validation: splitting our traversal loop into three, we train on two out of the three splits, evaluate performance on the third and average results across hold-out splits. Due to the computational demands of training models from scratch on each split, we train our medium resolution model (which is faster to train but has slightly worse performance than its higher resolution counterpart).

Our best model reduces average cross validation errors over the current state of the art by over $25\%$ in translational and $11\%$ in rotational error whilst running over 15x faster. 
Using this training paradigm we reduce the effective training data diversity by a third.
We attribute this to the slight reduction in performance in comparison to the results presented in Section \ref{sec:method-baseline}.
We theorise we could significantly boost performance by moving to our highest resolution model also.

\begin{table}[H]
\centering
\small
\centering
\begin{tabularx}{0.91\textwidth}{lccccc }
\toprule
                 & Resolution                                  & \multicolumn{2}{c}{Translational error ($\%$)} & \multicolumn{2}{c}{Rotational error (deg/m)} \\ 
 \textbf{Benchmarks}   & (m/pixel) & Mean & IQR &  Mean & IQR  \\ \cmidrule(lr){1-1} \cmidrule(lr){2-2} \cmidrule(lr){3-4} \cmidrule(lr){5-6} 

RO Cen Full Res  \cite{cen2018precise}      & 	0.0432  & 	6.3813          & 	4.6458 & 	0.0189          & 	0.0167 \\ \vspace{1mm} 
RO Cen Equiv.*  \cite{cen2018precise}       & 	0.1752  & 	\textit{3.6349} & 	3.3144 & 	\textit{0.0096} & 	0.0095 \\ \vspace{1mm} 
Raw Scan                                    & 	0.4     & 	8.4532          & 	8.0548 & 	0.0280          & 	0.0282 \\ \vspace{1mm}
Adapted Deep VO Cart \cite{li2018undeepvo}   & 	0.4     & 	11.531          & 	9.6539 & 	0.0336          & 	0.0307 \\ \vspace{1mm} 
Adapted Deep VO Polar \cite{li2018undeepvo} & 	        & 	14.446          & 	11.838 & 	0.0452          & 	0.0430 \\ \vspace{2mm} 
Visual Odometry \cite{WinstonChurchill}     & 	        & 	3.7824          & 	1.9884 & 	0.0103          & 	0.0072 \\ \vspace{1mm} 
\textbf{Ours}                               &           &                   &          &                    &          \\ \cmidrule(lr){1-1} \vspace{1mm}

Polar                                       & 	0.4     & 	2.8115 & 	2.4189 & 	0.0086 & 	0.0084 \\ \vspace{1mm} 
Cart                                        & 	0.4     & 	3.2756 & 	2.8213 & 	0.0104          & 	0.0100 \\ \vspace{1mm} 
Dual Polar                                  & 	0.4     & 	3.2359 & 	2.5760 & 	0.0098          & 	0.0091 \\ \vspace{1mm} 
Dual Cart                                   & 	0.4     & 	\textbf{2.7848} & 	2.2526 & 	\textbf{0.0085} & 	0.0080 \\
\bottomrule \\

\end{tabularx}

\caption{Spatial cross validation odometry estimation results. Our approach outperforms the benchmark (italics) in a large proportion of the experiments and we would expect a similar boost in performance to Section \ref{sec:method-baseline} by moving from our medium to highest resolution model. Our best performing model in terms odometry performance is marked in bold.
}
\label{table:cross_validation_odometry_results_average}

\end{table}
\vfill


\subsection{Additional Evaluation Examples}
\begin{figure}[H]
    
  \centering
  \includegraphics[width=1. \linewidth]{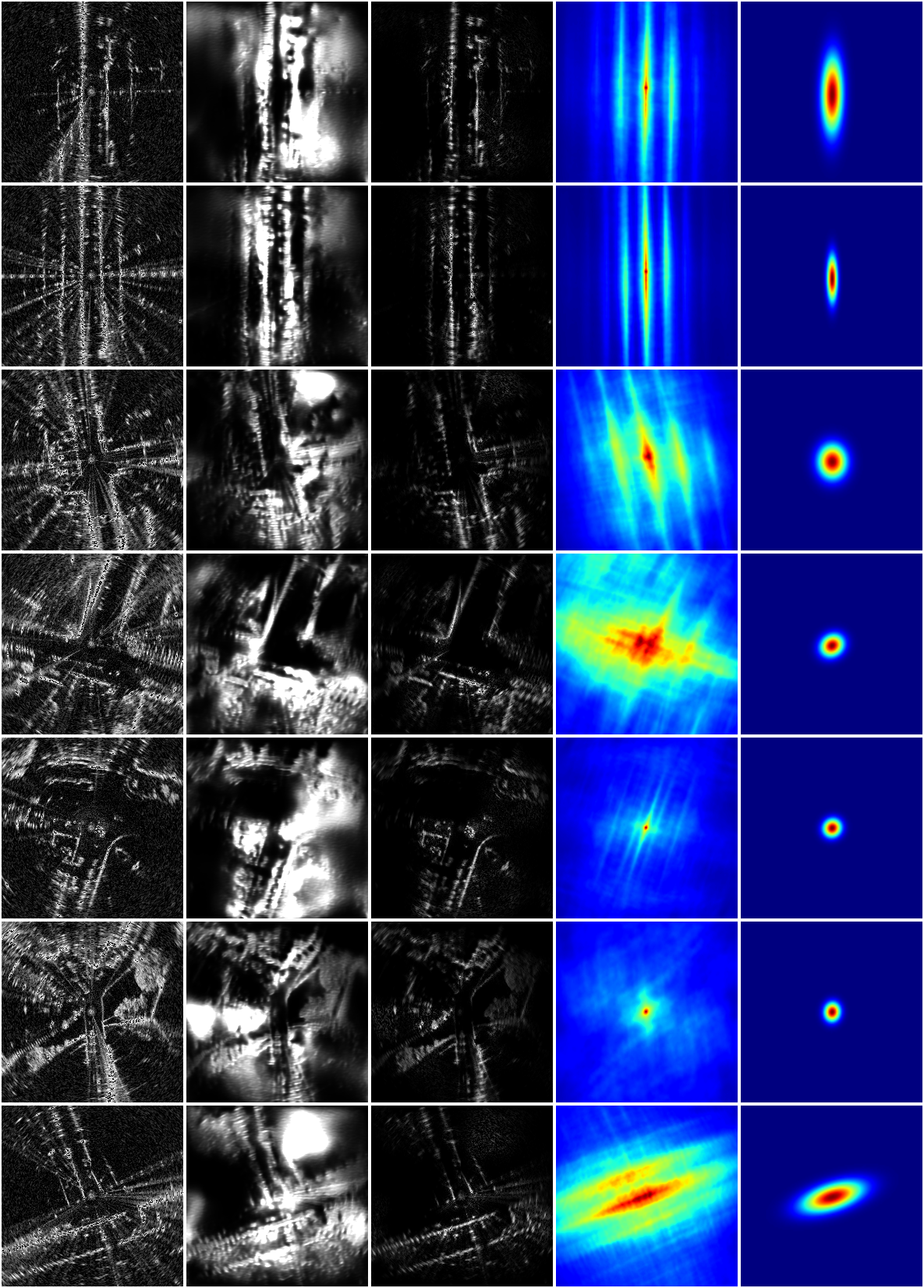}
  \begin{small}
  \begin{tabularx}{\textwidth}{YYYYY}
    Radar Input & Predicted Mask & Masked Features & Correlation & Covariance
  \end{tabularx}
  \vspace{-4mm}
  \end{small}
  \caption{Additional qualitative examples generated from our best performing model. The masks generated from our network filter out noise and distractor objects in the scene whilst preserving temporally consistent features such as walls, well suited for pose prediction. From left to right the raw Cartesian radar scan, the predicted network mask, the masked radar scan, the correlation volume and the fitted gaussian to the correlation volume after temperature weighted softmax.
  }
  \label{fig:further_examples}
\end{figure}


\newpage

\end{document}